# Multivariate Time series Anomaly Detection:
# A Framework of Hidden Markov Models

Jinbo Li[1] , Witold Pedrycz[2], and Iqbal Jamal[3]

[1] Department of Electrical & Computer Engineering, University of Alberta,
Edmonton, Alberta T6R 2V4 AB Canada ( jinbo@ualberta.ca )

[2] Department of Electrical & Computer Engineering, University of Alberta,
Edmonton, Alberta T6R 2V4 AB Canada

Department of Electrical and Computer Engineering, King Abdulaziz University,
Jeddah, 21589, Saudi Arabia,

Systems Research Institute, Polish Academy of Sciences,
Newelska 6, 01-447, Warsaw, Poland ( wpedrycz@ualberta.ca )

[3]AQL Management Consulting Inc.,
Edmonton, Alberta T6J 2R8, Canada ( iqbaljamal@aqlmc.com )

**Abstract**

In this study, we develop an approach to multivariate time series anomaly detection focused on the transformation of multivariate time series to univariate time series. Several transformation techniques involving Fuzzy C-Means (FCM) clustering and fuzzy integral are studied. In the sequel, a Hidden Markov Model (HMM), one of the commonly encountered statistical methods, is engaged here to detect anomalies in multivariate time series. We construct HMM-based anomaly detectors and in this context compare several transformation methods. A suite of experimental studies along with some comparative analysis is reported.

## 1. Introduction

Multivariate time series has become prevalent in a broad range of real-world applications such as weather data analysis and prediction [1], health care [2], finance [3-7] and others [8-11]. Anomaly detection, as an important class of problems in the analysis of multivariate time series, aims at finding abnormal or unexpected sequences. It has attracted significant attention in the recent decades. Whereas most existing research only focuses on anomaly detection in univariate time series, relatively less studies have been devoted to problems of detection of anomalies in multivariate time series. There are several factors that make anomaly detection of multivariate time series more complicated. The first difficulty arises because of the lack of a concise and operational anomaly definition [12]. Unusual points (exhibiting too high or too low values) and unexpected subsequences (e.g., shape changes) [13] appearing in univariate time series can be considered as anomaly. Unlike these definitions, multivariate techniques do not only deal with the abnormal values or subsequences in each time series but also investigate the relationships among these variables. For instance, when studying time series of records of photosynthesis of flowers/fruits grown outdoors, the measured values of photosynthesis (in summer) are evidently different from that (in winter) because temperature has a direct impact on photosynthesis of plants [14]. The stock price of one enterprise reflects its current performance and management. However, current macroeconomic variables, namely Gross Domestic Product (GDP), exchange rate, inflation rate and others, have more or less tangible impact on stock prices [15]. More research on this problem is reported in [16]. Secondly, the ubiquitous presence of noise will cause some errors in multivariate time series anomaly detection. The algorithm's robustness [12] against noise is expected to be useful in improving detection accuracy.

A number of methods have been introduced to find anomalies in multivariate time series. These methods can be grouped into three main categories [17]: (i) transformation (from high-dimensionality to single-dimensionality)-based methods, (ii) generative model-based methods and (iii) graph-based methods. Graph structures [12, 18, 19] are commonly exploited in methods of multivariate time series anomaly detection. In general, nodes of a graph represent subsequences or data points while the weights associated with the edges of the graph are aimed to capture similarity values of the corresponding nodes. Nevertheless, the potential limitation of these methods comes from the fact that more instances imply more time required to estimate the weights of its edges. Model-based methods [20] can also be implemented to detect anomalies of multivariate time series. In a certain sense, their benefits and drawbacks are associated with the input time series. In general, if the constructed model (e.g., state space models, vector models and fuzzy time series models [9, 21]) can predict a feature value based on past feature values accurately, it can assign relative accurate anomaly score to each subsequences or time point because the difference between the predicted and measured values is usually considered as anomaly score [22]. Unfortunately, without experts' knowledge of the system (temporal data), it is generally difficult to build accurately the pertinent model. In the case of transformation-based method, the easiest way is to calculate the average value over all variables at each time point and use the result to generate a new 'combined' sequence. The other commonly encountered transformation method relies on the use of the principal component analysis (PCA) that projects high dimensional time series into a low-dimensional sequence. However, during transformation process, information losses of multivariate time series become inevitable. In general, the transformation process consists of the two main steps, namely (i) information fusion, and (ii) discretization. There are a number of information aggregation/fusion strategies reported in the literature [23]. Its main objective is to combine multiple sources of information (e.g., multiple values present at the $t^{\text{th}}$ time instance) and provide a summarization of multiple variables of multivariate time series. For more studies about multivariate outliers detection, the reader may refer to [24, 25]. HMM, being regarded as one of the commonly used statistical model, can model the dynamic behavior of time series with a simple, yet powerful latent variable model [26]. This model has been successful in a wide range of applications such as credit ratings [27], fault diagnosis [28, 29], and others [30, 31]. Compared with the traditional Markov model, it includes states that are not directly visible and regarded as hidden states [26]. The observable states in the HMM follow a probability distribution (or an emission distribution) and depend on the hidden states. Based the observed states, the aforementioned feature of the HMM can provide a potential to determine if a subsequence or data point belongs to either an abnormal or normal category. In other words, for time series anomaly detection, two unobservable states (normal or abnormal) can explain the observed temporal sequence.

The main objective of this paper is to propose a HMM-based anomaly detectors for multivariate time series. Another objective of this study is to develop a HMM-based detector and demonstrate its performance in a range of practical applications. In this framework, we investigate some transformation methods and study their performance with respect to abilities to retain useful information (e.g., amplitude or amplitude change). If the combined sequence that represents multivariate time series can capture most useful information, a variety of existing univariate time series anomaly detection methods could be applied directly to such multivariate time series.

This paper is organized as follows. Section II is focused on a brief summary of HMM, fuzzy integrals and FCM. Section III introduces the proposed method. We elaborate on the performance of the method in Section IV. Finally, in Section V, we draw concluding comments.

## 2. Preliminaries

In this section, we first offer some concise summary of the HMM. Afterwards, we discuss some useful transformation methods aimed at the analysis of multivariate time series used in this paper.

### *2.1 HMM*

HMM can cope with time series that are generated by a certain Markov process. Two essential assumptions are made: (i) only the current states affect the next state, (ii), transition probabilities between the states do not vary over time (stationarity requirement). In particular, for each HMM, there are hidden/observed state sets and three probability matrices. Each hidden state emits one of the states that can be directly observed. The hidden state set $Q=\{q_1,q_2,\ldots,q_N\}$ comprises of $N$ possible hidden states and the observed state set $S=\{s_1,s_2,\ldots,s_M\}$ consists of $M$ possible observed states.

Let us assume an observed state sequence coming in the form $\boldsymbol{O}=o_1,o_2,\ldots,o_T$. To gain a clear understanding of the HMM, assume $\boldsymbol{I}=i_1,i_2,\ldots,i_T$ is the corresponding hidden state sequence of the above observed state sequence. For each HMM, it can be defined as follows.

$$\lambda=(\boldsymbol{A},\boldsymbol{B},\boldsymbol{\Pi}) \tag{1}$$

$\boldsymbol{A}=[a_{ij}]$ $(1\le i,j\le N)$ and $\boldsymbol{B}=[b_{ik}]$ $(1\le i\le N, 1\le k\le M)$ are the state transition matrix and emission matrix, respectively. $a_{ij}=P[i_{t+1}=q_j \mid i_t=q_i]$ denotes the probability that the state $q_i$ in $t^{\text{th}}$ time moves to $q_j$ in $(t+1)^{th}$ time while $b_{ik}=P[o_t=s_k \mid i_t=q_i]$ stands for the probability of observed state $s_k$ in the $t^{th}$ time when the hidden state is $q_i$ at this time moment. $\boldsymbol{\Pi}=[\Pi_i](1\le i\le N)$ is initial vector. $\Pi_i=P[i_1=q_i]$ denotes the probability of the hidden state $q_i$ occurring in the $i^{th}$ time moment

In general, HMM deals with the three standard problems that arise in various applications:

- Given a HMM model, $\lambda = (\boldsymbol{A}, \boldsymbol{B}, \boldsymbol{\Pi})$ and a observed sequence $\boldsymbol{O} = o_1, o_2, \ldots, o_T$, calculate the probability $P(\boldsymbol{O} \mid \lambda)$ that the observed sequence has been produced by this HMM $\lambda = (\boldsymbol{A}, \boldsymbol{B}, \boldsymbol{\Pi})$.
- Given an observed sequence $\boldsymbol{O} = o_1, o_2, \ldots, o_T$ , estimate the parameters of the HMM model $\lambda = (\boldsymbol{A}, \boldsymbol{B}, \boldsymbol{\Pi})$ that maximize the probability $P(\boldsymbol{O} \mid \lambda)$ of observations given the model.
- Given a HMM model $\lambda = (\boldsymbol{A}, \boldsymbol{B}, \boldsymbol{\Pi})$ and a observed sequence $\boldsymbol{O} = o_1, o_2, \ldots, o_T$, decide the most likely state sequence $\boldsymbol{I}$

The Viterbi algorithm, realizing an algorithm of dynamic programming algorithm, estimates the most probable state sequence [32]. It can determine the optimal hidden state sequence $\boldsymbol{I} = i_1, i_2, \ldots, i_T$ based on the HMM model $\lambda = (\boldsymbol{A}, \boldsymbol{B}, \boldsymbol{\Pi})$ and the given observed state sequence $\boldsymbol{O} = o_1, o_2, \ldots, o_T$.

Consider that $\delta_t(i)$ stands for the probability of state $i$ at $t^{th}$ time moment defined as follows

$$\delta_t(i) = \max_{i_1, i_2, \ldots, i_{t-1}} P(i_t = i, i_{t-1}, \ldots, i_1, o_t, o_{t-1}, \ldots, o_1 | \lambda) = \max_{1 \le j \le N} \left( \delta_t(j) a_{ji} \right) b_{io_{t+1}}$$

The detailed algorithm comes as the following sequence of steps.

Initialization:

$$\delta_1(i) = \pi_i b_{io_1} \tag{2}$$

Recursion:

$$\delta_{t+1}(i) = \max_{i_1, i_2, \ldots, i_t} P(i_{t+1} = i, i_t, \ldots, i_1, o_{t+1}, o_t, \ldots, o_1 | \lambda) = \max_{1 \le j \le N} \left( \delta_t(j) a_{ji} \right) b_{io_{t+1}} \tag{3}$$

Termination:

$$P^* = \max_{1 \le j \le N} \delta_T(i) \tag{4}$$

As an illustrative example, Figure 1 shows a simple example of the HMM when the numbers of hidden states and observed states are 2 and 3, respectively.

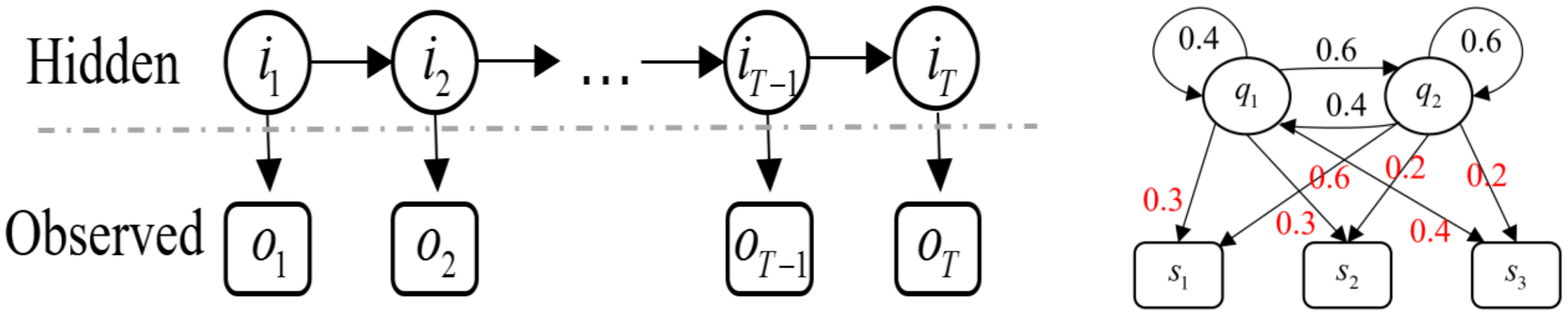

(a) (b)

Figure 1 Illustrative example of HMM (red: emission probabilities; black: transition probabilities)

*2.2 Multivariate time series transformation methods*

In practice, the values of multivariate time series are collected using different sensors. There are a number of information-retaining methods for transforming multivariate time series into an observed sequence. Here, we focus on FCM clustering methods and fuzzy integral methods, which were found useful in many applications [33].

*2.2.1 FCM Algorithm*

A sound alternative to transform multivariate time series to an observed sequence is to use FCM clustering algorithm [34, 35]. Given a multivariate time series $\boldsymbol{X} = \boldsymbol{x}_1, \boldsymbol{x}_2, \ldots, \boldsymbol{x}_T$ of length $T$ , the objective function $Q$ used in the FCM is defined in the following way

$$Q = \sum_{i=1}^{c}\sum_{j=1}^{T} u_{ij}^{m} d^2\left(\boldsymbol{x}_{j,}\boldsymbol{v}_i\right) \tag{5}$$

Here $c$ stands for the number of clusters and $m(m>1)$ denotes the fuzzification coefficient. $U=[u_{ij}]$ and $\boldsymbol{v}_i$ are the partition matrix and the $i^{th}$ prototype, respectively. $d^2\left(\boldsymbol{x}_{j,}\boldsymbol{v}_i\right)$ (as well as $\|.\|^2$) stands for the Euclidean distance (or its generalization) between $\boldsymbol{x}_{j,}$ and the prototype $\boldsymbol{v}_i$. The partition matrix and cluster centers (prototypes) are calculated iteratively as follows

$$\boldsymbol{v}_i = \frac{\sum_{j=1}^{T} u_{ij}^{m}\boldsymbol{x}_j}{\sum_{j=1}^{T} u_{ij}^{m}} \tag{6}$$

$$u_{ij} = \frac{1}{\sum_{l=1}^{c}\left(\frac{\|\boldsymbol{v}_i - \boldsymbol{x}_j\|}{\|\boldsymbol{v}_l - \boldsymbol{x}_j\|}\right)^{2/(m-1)}} \tag{7}$$

The sequence of iterations is carried out to realize the minimization of the objective function. Then, on a basis of the partition matrix generated by the FCM, each $\boldsymbol{x}_j$ belongs to the cluster to which it exhibits the highest membership degree.

### 2.2.2 *Fuzzy measures and fuzzy integrals*

Fuzzy integrals can combine different sources of uncertain information [36] and have been widely applied to a variety of fields, such as decision making [37], pattern recognition [38],supplier evaluation [39], gaze control of robotics [40], etc. [41]. Fuzzy integral is calculated with respect to a fuzzy measure that can capture the relationship among different variables. Let us recall that by a fuzzy measure we mean a set function $g$ that satisfies the following set of conditions

Boundary conditions:

$$g(\phi)=0 \quad g(\boldsymbol{X})=1 \tag{8}$$

Monotonicity:

$$\text{IF} \quad A\subset B\big(A,B\in g(X)\big), \quad \text{THEN} \quad g(A)\le g(B) \tag{9}$$

Continuity:

If $\{A_n\},(1\le n\le\infty)$ is a monotone sequence of measurable sets, then

$$\lim_{n\to\infty} g(A_n)=g\Big(\lim_{n\to\infty} A_n\Big) \tag{10}$$

Based on the above definition, Sugeno developed a certain type of fuzzy measures, namely $\lambda$-fuzzy measure [42]. Here the union of two disjoint sets $A$ and $B$ is determined as follows.

$$g(A\cup B)=g(A)+g(B)+\lambda g(A)g(B) \tag{11}$$

Based on the normalization condition, the parameter of $\lambda$ describes a level of interaction between the two disjoint sets and is greater than -1. The determination of its value comes as a result of solution to the following polynomial equation

$$\lambda+1=\prod_{i=1}^{n}(1+\lambda g_i) \quad \lambda>-1 \tag{12}$$

where $\lambda$ models several types of interaction: excitatory for its positive values, inhibitory for the negative values. The fuzzy measure is additive (no interaction) when $\lambda$ =0. In what follows, we recall a concept of the fuzzy integrals.

*Sugeno fuzzy integral*

Let $g$ be a fuzzy measure. Let $h$ be a function: $X \rightarrow [0,1]$. The Sugeno fuzzy integral of $h$ with respect to the fuzzy measure $g$ is calculated in the following form

$$\int_A h(x) \circ g = \sup_{\alpha \in [0,1]} \left[ \min\left(\alpha, g\left(A \cap H_\alpha\right)\right) \right] \tag{13}$$

Where $H_\alpha = \{x \mid h(x) \geq \alpha\}$ is an $\alpha$-cut of this function.

*Choquet fuzzy integral*

Let $g$ be a fuzzy measure. As before $h$: $X \rightarrow [0,1]$ . The Choquet fuzzy integral of $h$ with respect to $g$ is expressed in the following form

$$\int_A h(x) \circ g = \sum_{i=1}^{n} \left[ h(x_i) - h(x_{i-1}) \right] g(A_i) \tag{14}$$

Here $g(A_i) = g_i + g(A_{i-1}) + \lambda g_i g(A_{i-1})$.

For Sugeno Integral and Choquet Integral determined with respect to the $\lambda$-fuzzy measure, the calculation of integral only requires information about fuzzy density [43] $g_i$. Higher values of $g_i$ indicate that the $i$th feature is increasingly essential. As an illustrative example, we consider a single multivariate time series involving three variables reported at a certain time moment and recording measurement values of three sensors. Here the quality of information from each sensor can be regarded as the value of the fuzzy density. Higher values of the fuzzy density $g_i$ indicate more essential entries at this time moment. The values of multivariate time series are arranged in a vector form

$$h = [0.7 \;\; 0.4 \;\; 0.3]$$

The corresponding vector of the fuzzy densities assumes the following entries (those values can be estimated by experts or derived on a basis of some training data).

$$g = [0.21 \;\; 0.35 \;\; 0.05]$$

Following the above definitions, the results of Sugeno fuzzy integral and Choquet fuzzy integral are equal to 0.4400 and 0.7889, respectively.

## 3. Problem formulation and the proposed solution

Let us assume a multivariate time series $\boldsymbol{X} = \boldsymbol{x}_1, \boldsymbol{x}_2, \ldots, \boldsymbol{x}_{T+T'}$ of length $T + T'$. $T$ and $T'$ are the lengths of training and testing time series, respectively. After applying z-score normalization [44], we run FCM, and then determine both Sugeno integral and Choquest integral to produce training (and testing) (observed) state sequences $\boldsymbol{O} = o_1, o_2, \ldots, o_T$ (and $\boldsymbol{O'} = o'_1, o'_2, \ldots, o'_{T'}$ ) .Compared with the FCM, an additional step, namely mapping/vector discretion (from continuous to discrete), is necessary for construct fuzzy integral based detectors.

In the construction of the HMM, we use a labeled training state sequence coming in form of data-label pairs $(o_k, i_k)$, $k = 1, 2, \ldots, T$ where $o_k$ is one-dimensional observed state and $k_k$ stands for its label (normal or abnormal). The temporally ordered labels are regarded as a hidden state sequence of the HMM.

Considering the time series anomaly detection problem, for HMM, the number of hidden states is equal to 2, which correspond to normal or abnormal state. Thus, the initial vector $\boldsymbol{\Pi}$ , state transition matrix $\boldsymbol{A}$ , and emission matrix $\boldsymbol{B}$ are expressed as follows.

$$\boldsymbol{\Pi} = \left[ \Pi_1(normal) \quad \Pi_2(abnormal) \right]$$

$$\boldsymbol{A} = \begin{bmatrix} a_{11}(normal \to normal) & a_{12}(normal \to abnormal) \\ a_{21}(abnormal \to normal) & a_{22}(abnormal \to abnormal) \end{bmatrix}$$

$$\boldsymbol{B} = \begin{bmatrix} b_{11} & \cdots & b_{1M} \\ b_{21} & \cdots & b_{2M} \end{bmatrix}$$

Next we calculate the state transition matrix and the emission as follows.

$$\Pi_i = \frac{|q_i|}{\sum q} \tag{15}$$

$$a_{ij} = \frac{|q_{ij}|}{\sum_{j=1}^{N} |q_{ij}|} \tag{16}$$

$$b_{ik} = \frac{|o_{il}|}{\sum_{l=1}^{M} |o_{il}|} \tag{17}$$

Once the parameters of HMM have been determined, Viterbi algorithm is then used to determine the label of testing observed state sequence $\boldsymbol{O'} = o'_1, o'_2, \ldots, o'_{T'}$. The most likely state sequence is $\boldsymbol{I}' = i'_1, i'_2, \ldots, i'_{T'}$ . Let us highlight the essence of the proposed methods as shown in Figure 2; we point at the two key methodological steps encountered there, namely a transformation from multivariate time series to univariate time series followed by HMM-based detection. After application of the z-score normalization, we invoke different transformation methods, namely FCM, Sugeno integral and Choquet integral, which implement the transformation. Subsequently, we estimate essential parameters of the HMM by using the labels of the training set. More specifically, the collected normal and abnormal time points are considered to estimate the emission and transition probabilities of the HMM. After training the HMM, we apply the Viterbi algorithm to test the observed state sequence and compute the most likely hidden state sequence consisting of the two states ( normal and abnormal).

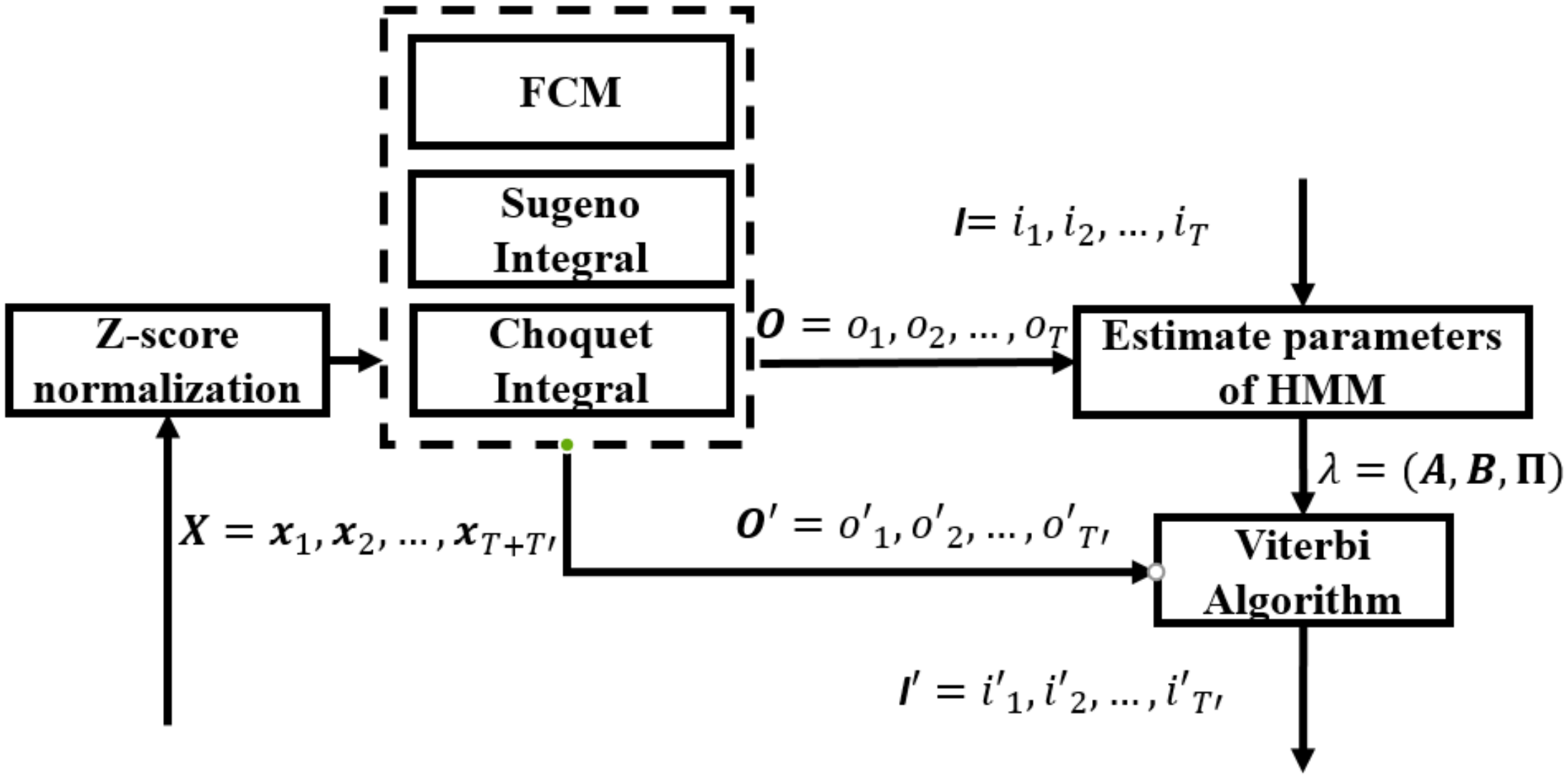

Figure 2. Overall processing realized by the anomaly detector

## 4. Experimental Studies

In this section, we report on a series of numeric examples illustrating how the amplitude anomalies in multivariate time series are detected. Both synthetic data and the publicly available datasets with artificial anomalies are considered.

### 4.1. Synthetic data

The multivariate time series is generated in the form of sine and cosine functions of different frequencies, see Figure 2. The length of the series is equal to 2,000 samples and there are some visible changes at different time points of each variable of multivariate time series. Gaussian noise (with the zero mean and unit standard deviation) is added to each variable of the original multivariate time series to increase the difficulty of detecting the anomalies and make the data more realistic. These artificial anomalies are generated by randomly picking some time points and increasing their amplitude by multiplying them by a random value located in the interval [0, 3].

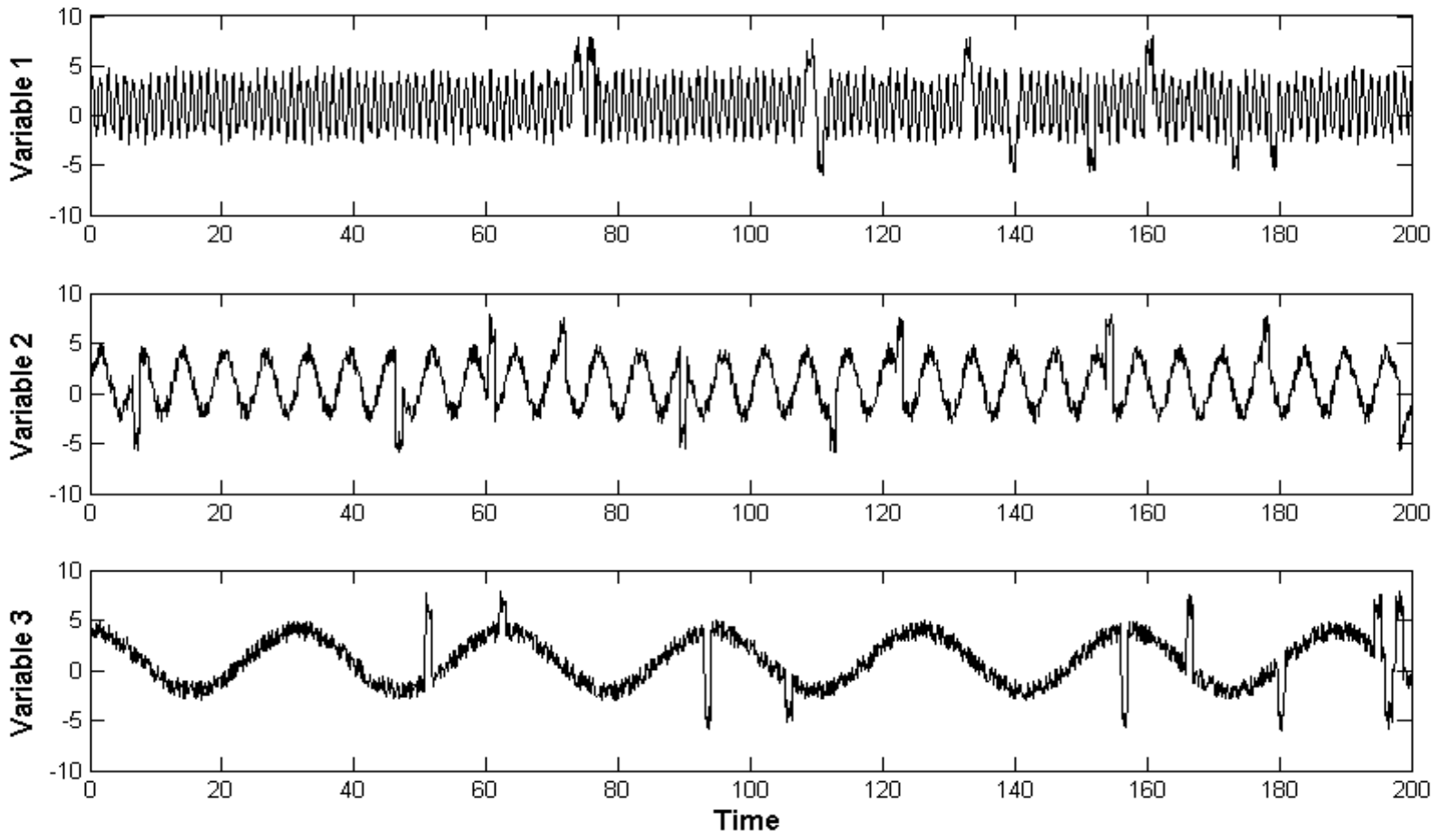

Figure 3. Synthetic multivariate time series

Two data sets, one covering time points from 1 to 140 (treated as a training set) and another one covering time points from 140.1 to 200 (testing set), have been considered in this experiment. For comparison, PCA is also exploited to transform multivariate time series to an observed sequence. As the first component of the PCA transformed data captures the most information about the data [45], it would be possible to use only this component (the one with the highest eigenvalue) as a new 'combined' sequence, obtaining a transformation from multivariate time series to univariate time series.

To cluster the multivariate time series, there are two essential parameters of the FCM, namely a fuzzification coefficient and the number of clusters. Here we vary the values of the fuzzification coefficients ranging from 1.1 to 2.9 with a step of 0.1 while the number of clusters is taken from 2 to 198. For discretization, different values of the number of observed states located in the range [2, 80] have been considered leading to the optimal value of this parameter.

Figure 3 displays the experimental results produced by different methods. To make results more readable, for each detector, its objective (or quantitative) evaluation on this dataset have been reported to evaluate its performance. When *TP*, *FP*, *TN* and *FN* are the number of normal time points correctly detected as normal (True Positives), the number of abnormal time points that are detected as normal (False Positives), the number of abnormal time points that are detected as abnormal (True Negatives) and the number of normal time points that are detected as abnormal (False Negatives), Accuracy, sensitivity, specificity and F-measure are defined as the following expressions, where are the objective (or quantitative) evaluation included in our experiments. Table 1 displays the confusion matrices produced by different methods

$$\text{Accuracy} = \frac{TN+TP}{TN+FP+FN+TP} \tag{18}$$

$$\text{Sensitivity}=\frac{TP}{TP+FN} \tag{19}$$

$$\text{Specificity}=\frac{TN}{TN+FP} \tag{20}$$

$$\text{F-measure}=\frac{2\times\text{Precision}\times\text{Recall}}{\text{Precision}+\text{Recall}} \tag{21}$$

where

$$\text{Precision} = \frac{TP}{TP+FP} \quad \text{and} \quad \text{Recall}=\frac{TP}{TP+FN} \tag{22}$$

Table 1 Confusion matrix produced by different methods

| PCA+HMM | | | FCM+HMM | | |
|---|---|---|---|---|---|
| | p | n | | p | n |
| Y | 459 | 90 | Y | 464 | 94 |
| N | 16 | 35 | N | 11 | 31 |
| Sugeno integral+HMM | | | Choquet integral+HMM | | |
| | p | n | | p | n |
| Y | 473 | 65 | Y | 426 | 53 |
| N | 2 | 60 | N | 49 | 72 |

As shown in Table 2, the proposed FCM+HMM-based detector has achieved higher accuracy in comparison with the accuracy obtained when using other detectors. The accuracy improvement of FCM, Sugeno fuzzy integral, Choquet fuzzy integral based detectors vis-a-vis the generic PCA-based detector is in the range 7-9 %.

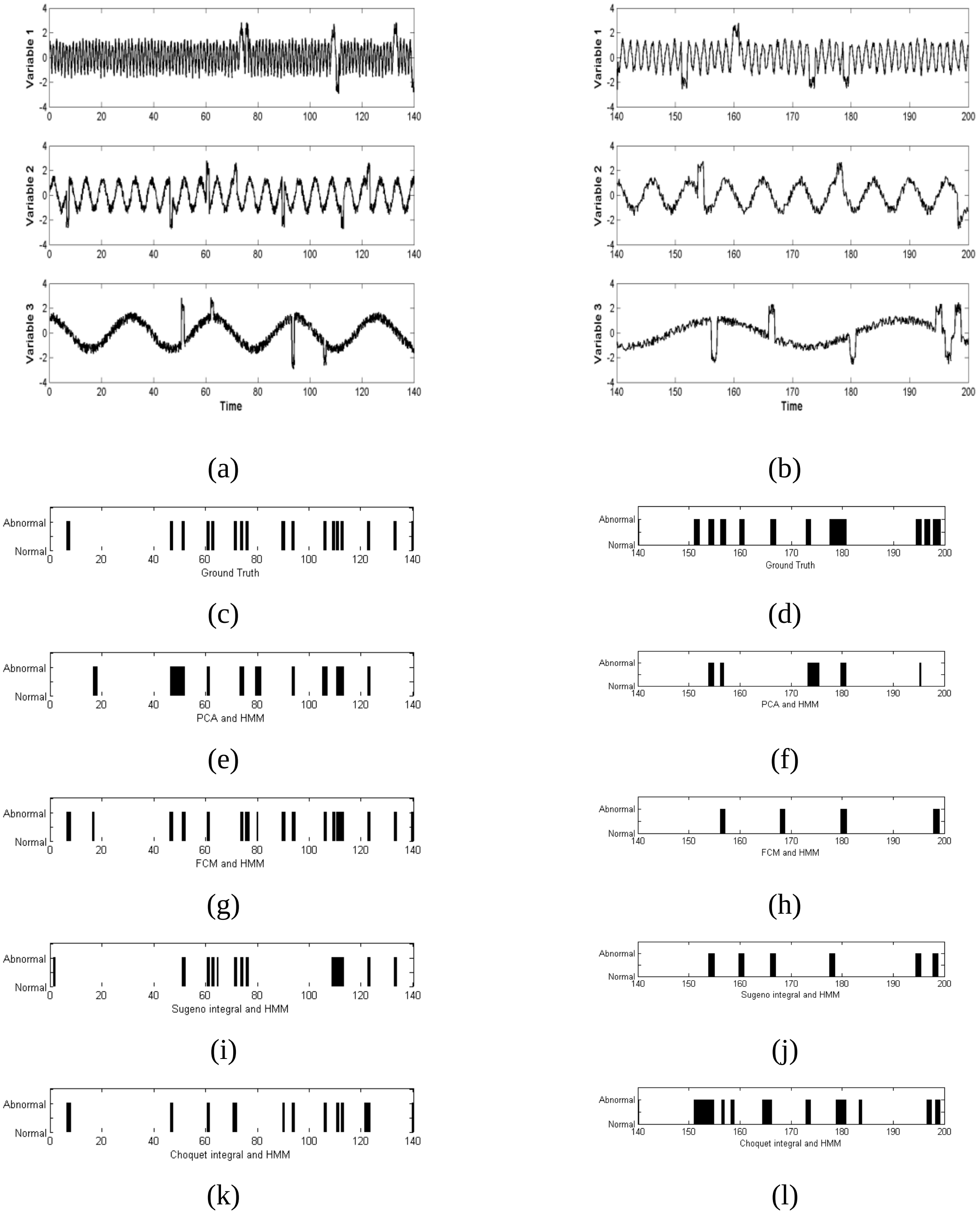

Figure 4 Synthetic multivariate time series: (a) training set, (b) testing set, (c) Ground truth of training set, (d) Ground truth of testing set, (e) Experimental results of PCA + HMM (training set), (f) Experimental results of PCA+HMM (testing set), (g) Experimental results of FCM + HMM (training set), (h) Experimental results of FCM + HMM (testing set), (i) Experimental results of Sugeno integral + HMM (training set), (j) Experimental results of Sugeno integral + HMM (testing set), (k) Experimental results of Choquet integral + HMM (training set), (l) Experimental results of Choquet integral + HMM (testing set).

Table 2 Experimental results obtained for synthetic multivariate time series

| Training Set | Accuracy | F-measure | Sensitivity | Specificity |
|---|---|---|---|---|
| PCA+HMM | 0.8764 | 0.9294 | 0.9245 | 0.5266 |
| FCM+HMM | 0.9550 | 0.9742 | 0.9651 | 0.8817 |
| Sugeno_Integral+HMM | 0.9350 | 0.9634 | 0.9740 | 0.6509 |
| Choquet_Integral+HMM | 0.9393 | 0.966 | 0.9838 | 0.6154 |
| Testing Set | Accuracy | *F*-measure | Sensitivity | Specificity |
| PCA+HMM | 0.8233 | 0.8964 | 0.9663 | 0.2800 |
| FCM+HMM | 0.8250 | 0.8984 | 0.9768 | 0.2480 |
| Sugeno_Integral+HMM | 0.8883 | 0.9338 | 0.9958 | 0.4800 |
| Choquet_Integral+HMM | 0.8300 | 0.893 | 0.8968 | 0.5760 |

To quantify the obtained optimal number of clusters and the value of the fuzzifiction coefficient, Figure 4 shows the corresponding accuracy when considering different values of these parameters. It is evident that the increase of the number of clusters will affect the performance of the detector significantly. The fuzzification coefficient exhibits some impact on the performance of the detector. Note that here HMM would fail due to the unknown external observed states that do not appear in training set. In other words, for adding new observations, re-training different HMM for new observations is anticipated.

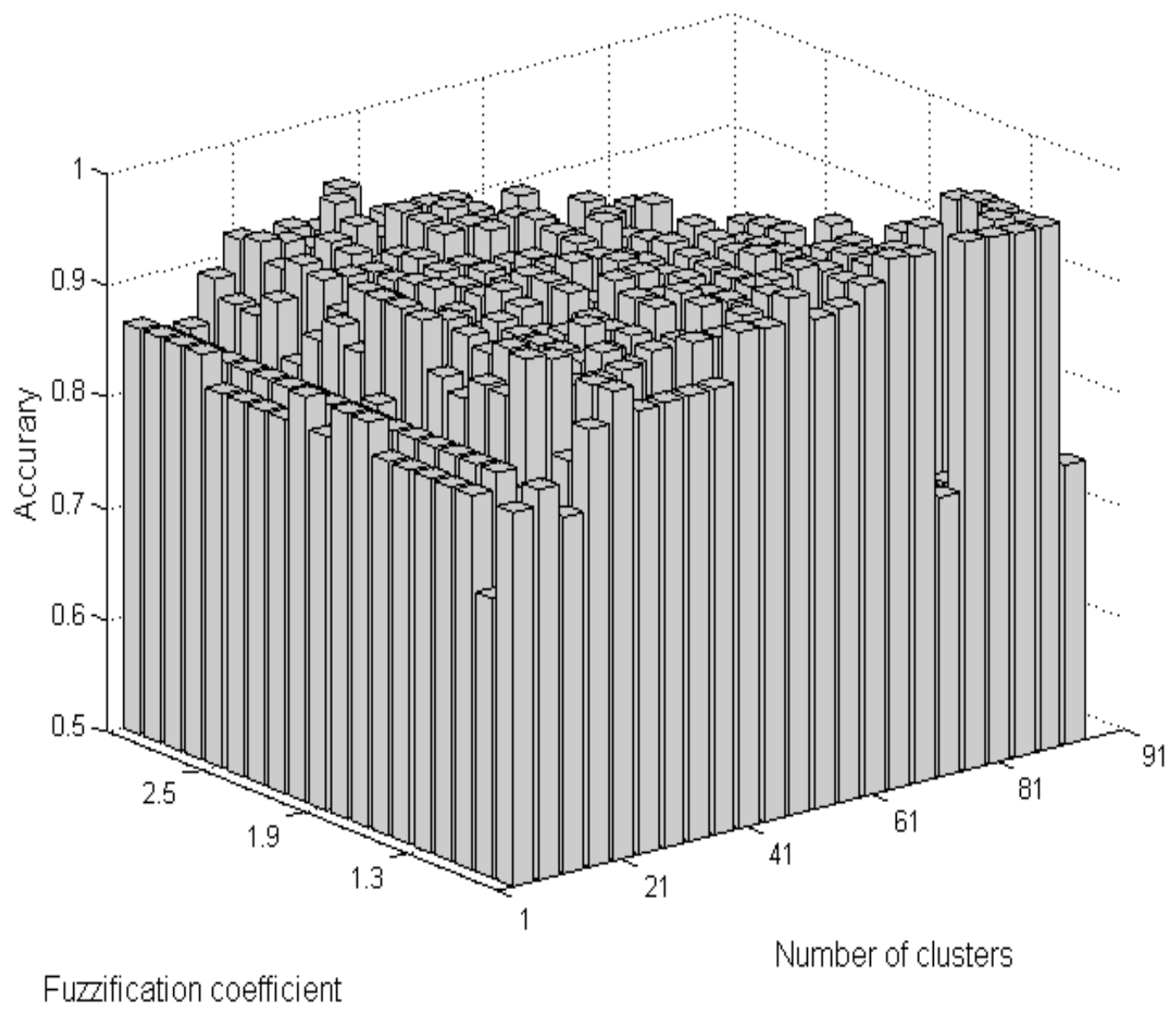

Figure 5 Performance comparison reported for various values of the fuzzification coefficient and the number of clusters

### 4.2. Publicly available datasets

In this subsection, we report on a variety of experiments on real world multivariate time series from different repositories such as UCI machine learn repository [47] and DataMarket [46]. The parameter setting is in the same way as presented for the synthetic data.

Data Set #1 [U.S. Dollar Exchange Rate]: The historical intraday data (per day except for holidays and regular weekends) for three currencies (the US dollar exchange rate versus the Dutch guilder, the French franc and the German mark) in the period January 03, 1989 to December 31, 1998: 1) Dutch guilder (NLG); 2) French franc (FF); 3) German mark (DEM).

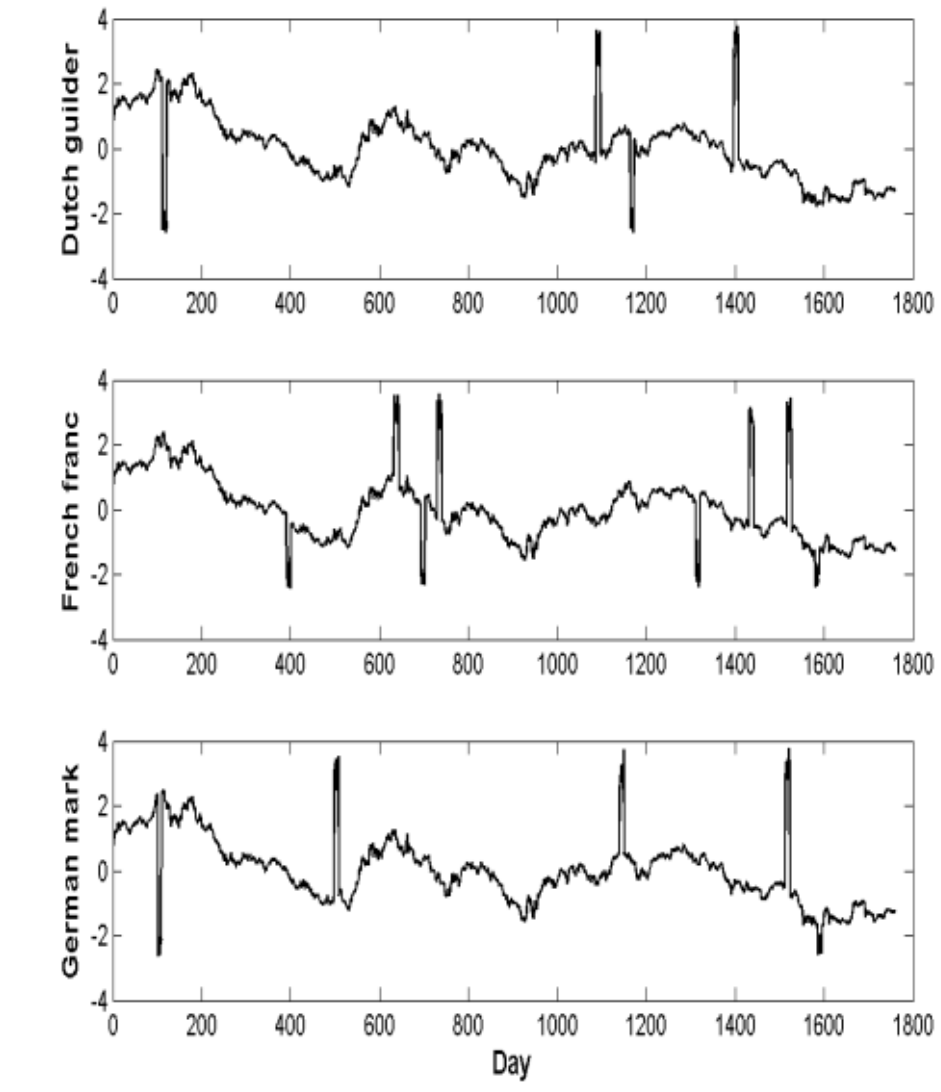

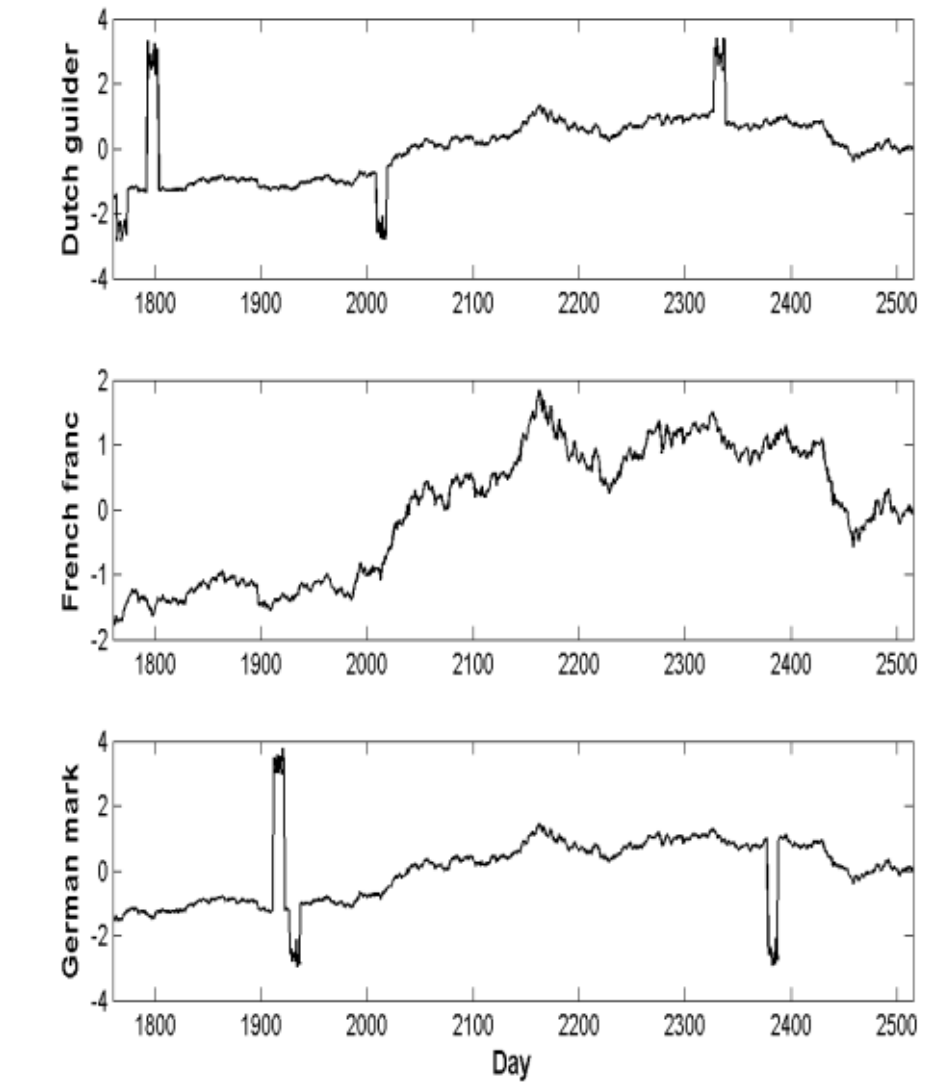

(a) (b)

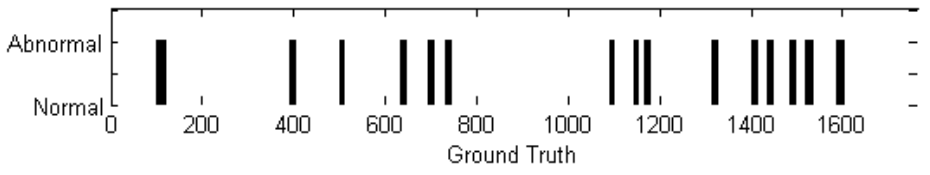

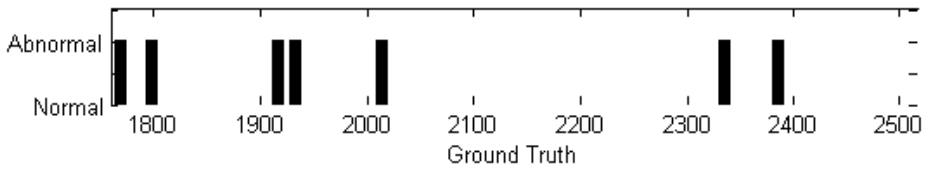

(c) (d)

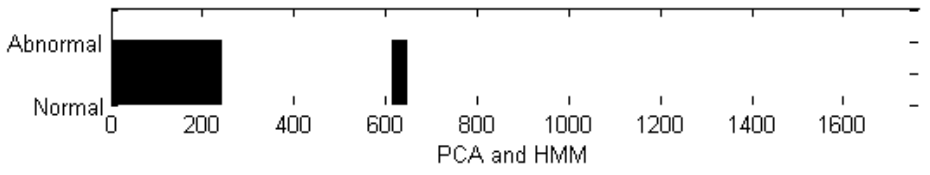

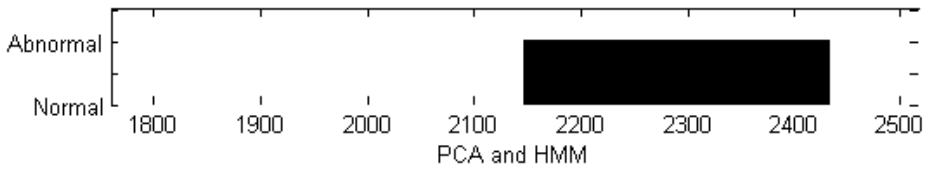

(e) (f)

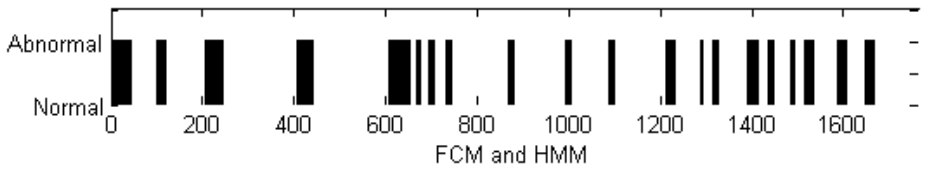

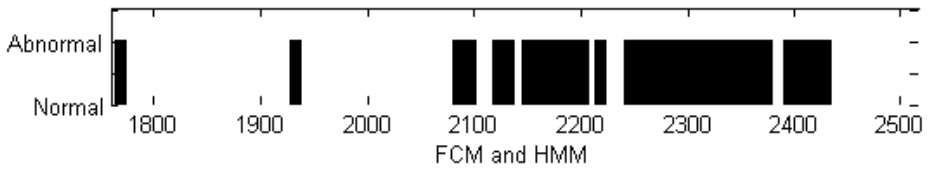

(g) (h)

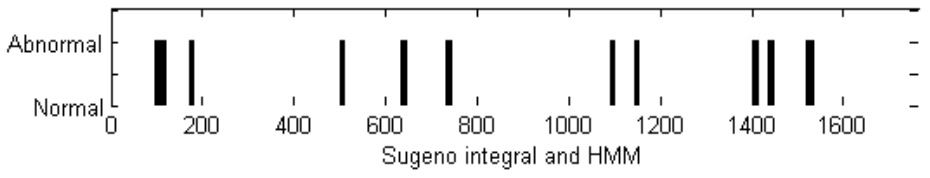

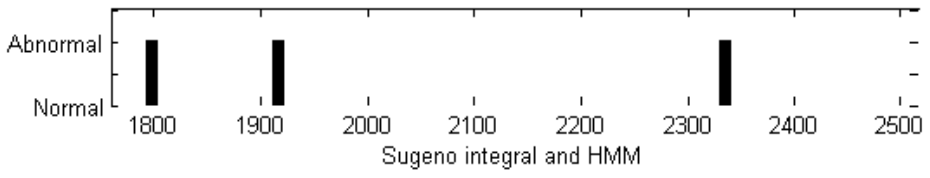

(i) (j)

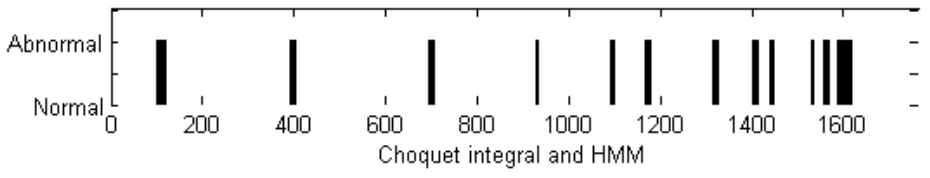

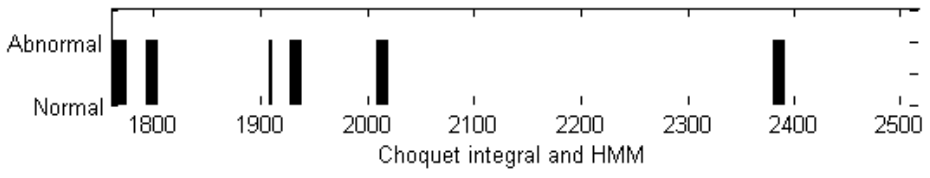

(k) (l)

Figure 6 U.S. Dollar Exchange Rate Dataset: (a) training set, (b) test set, (c) Ground truth of training set, (d) Ground truth of testing set, (e) Experimental results of PCA + HMM (training set), (f) Experimental results of PCA+HMM (testing set), (g) Experimental results of FCM + HMM (training set), (h) Experimental results of FCM + HMM (testing set), (i) Experimental results of Sugeno integral + HMM (training set), (j) Experimental results of Sugeno integral + HMM (testing set), (k) Experimental results for Choquet integral + HMM (training set), (l)

Experimental results of Choquet integral + HMM (testing set).

Table 3 Experimental results of U.S. Dollar Exchange Rate Dataset

| Training Set | Accuracy | Sensitivity | Specificity | F-measure |
|---|---|---|---|---|
| PCA+HMM | 0.7864 | 0.8516 | 0.1765 | 0.878 |
| FCM+HMM | 0.8386 | 0.8478 | 0.7529 | 0.9046 |
| Sugeno_Integral+HMM | 0.9523 | 0.9899 | 0.6000 | 0.974 |
| Choquet_Integral+HMM | 0.9483 | 0.9818 | 0.6353 | 0.9716 |
| **Testing Set** | **Accuracy** | **Sensitivity** | **Specificity** | **F-measure** |
| PCA+HMM | 0.5828 | 0.6131 | 0.2857 | 0.7272 |
| FCM+HMM | 0.5748 | 0.5898 | 0.4286 | 0.7156 |
| Sugeno_Integral+HMM | 0.9470 | 1.0000 | 0.4286 | 0.9716 |
| Choquet_Integral+HMM | 0.9669 | 0.9927 | 0.7143 | 0.982 |

Data Set #2 [EEG Eye State Dataset]: Three major EEG (eletroencephalogram) measurements (at a sampling frequency of 128 samples per second) acquired using the Emotiv EEG Neuroheadset: 1) AF3 (Intermediate between Fp and F); 2) F7 (Frontal left Hemisphere); 3) FC5 (Between F and C left Hemisphere).

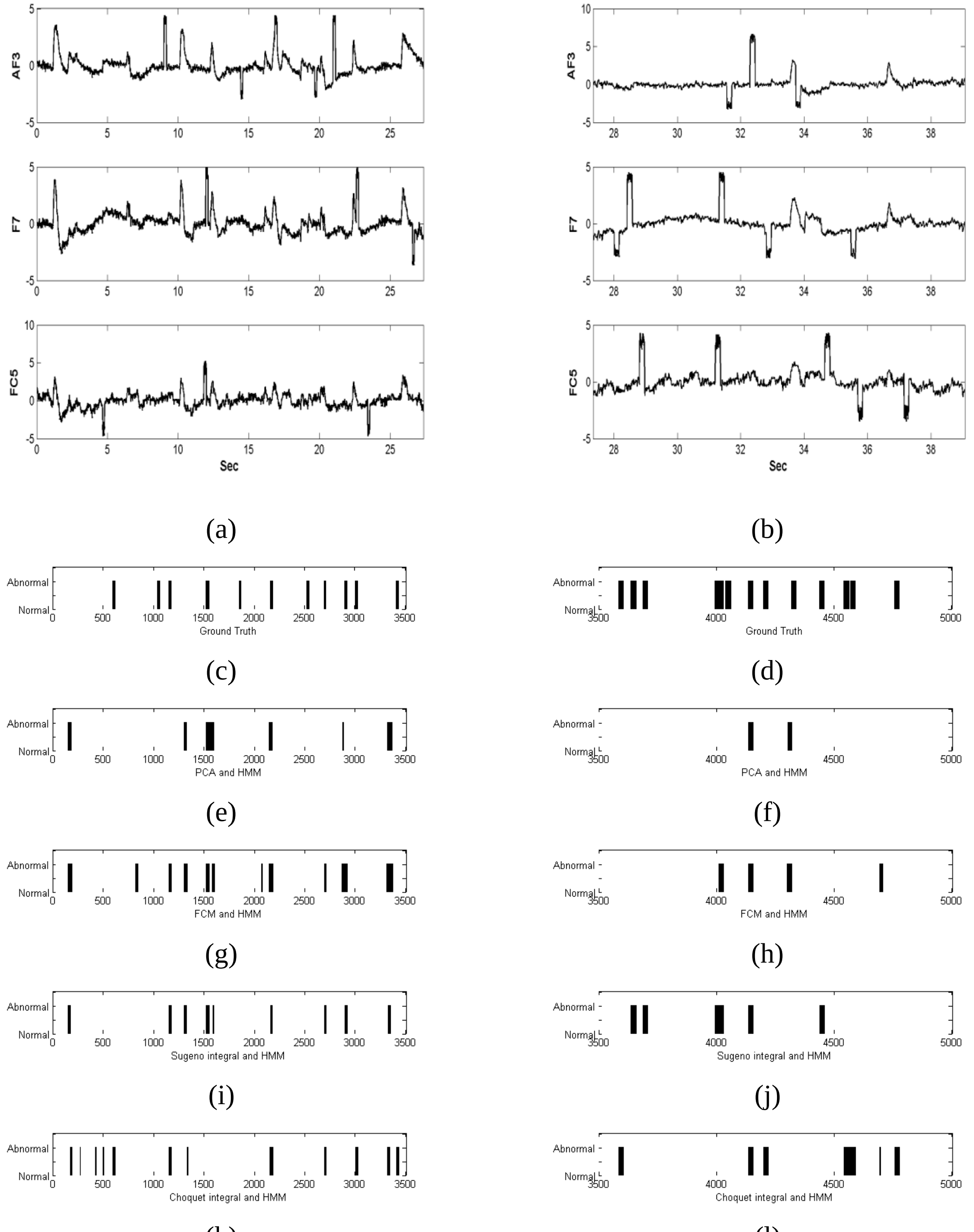

Figure 7 EEG Eye State Dataset: (a) training set, (b) test set, (c) Ground truth of training set, (d) Ground truth of testing set, (e) Experimental results of PCA + HMM (training set), (f) Experimental results of PCA+HMM (testing set), (g) Experimental results of FCM + HMM (training set), (h) Experimental results of FCM + HMM (testing set), (i) Experimental results of Sugeno integral + HMM (training set), (j) Experimental results of Sugeno integral + HMM (testing set), (k) Experimental results of Choquet integral + HMM (training set), (l) Experimental

results of Choquet integral + HMM (testing set).

Table 4 Experimental results of EEG Eye State Dataset

| Training Set | Accuracy | Sensitivity | Specificity | F-measure |
|---|---|---|---|---|
| PCA+HMM | 0.8997 | 0.9488 | 0.2203 | 0.9464 |
| FCM+HMM | 0.9003 | 0.9305 | 0.4831 | 0.9456 |
| Sugeno_Integral+HMM | 0.9454 | 0.9795 | 0.4746 | 0.971 |
| Choquet_Integral+HMM | 0.9477 | 0.9822 | 0.4703 | 0.9722 |
| **Testing Set** | **Accuracy** | **Sensitivity** | **Specificity** | **F-measure** |
| PCA+HMM | 0.8320 | 0.9879 | 0.0742 | 0.907 |
| FCM+HMM | 0.8340 | 0.9735 | 0.1563 | 0.9068 |
| Sugeno_Integral+HMM | 0.9067 | 1.0000 | 0.4531 | 0.9468 |
| Choquet_Integral+HMM | 0.9013 | 0.9904 | 0.4688 | 0.9434 |

Data Set #3 [Air Quality Dataset]: Three major chemical sensors (related to hourly average concentrations for Total Nitrogen Oxide, Nitrogen Dioxide and Ozone) produced by an Air Quality Chemical Multi-sensor Device that placed in a polluted field of an Italian city in the period March 10, 2004 to June 2, 2004: 1) PT08S3 ($NO_x$); 2) PT08S4 ($NO_2$); 3) PT08S5 ($O_3$).

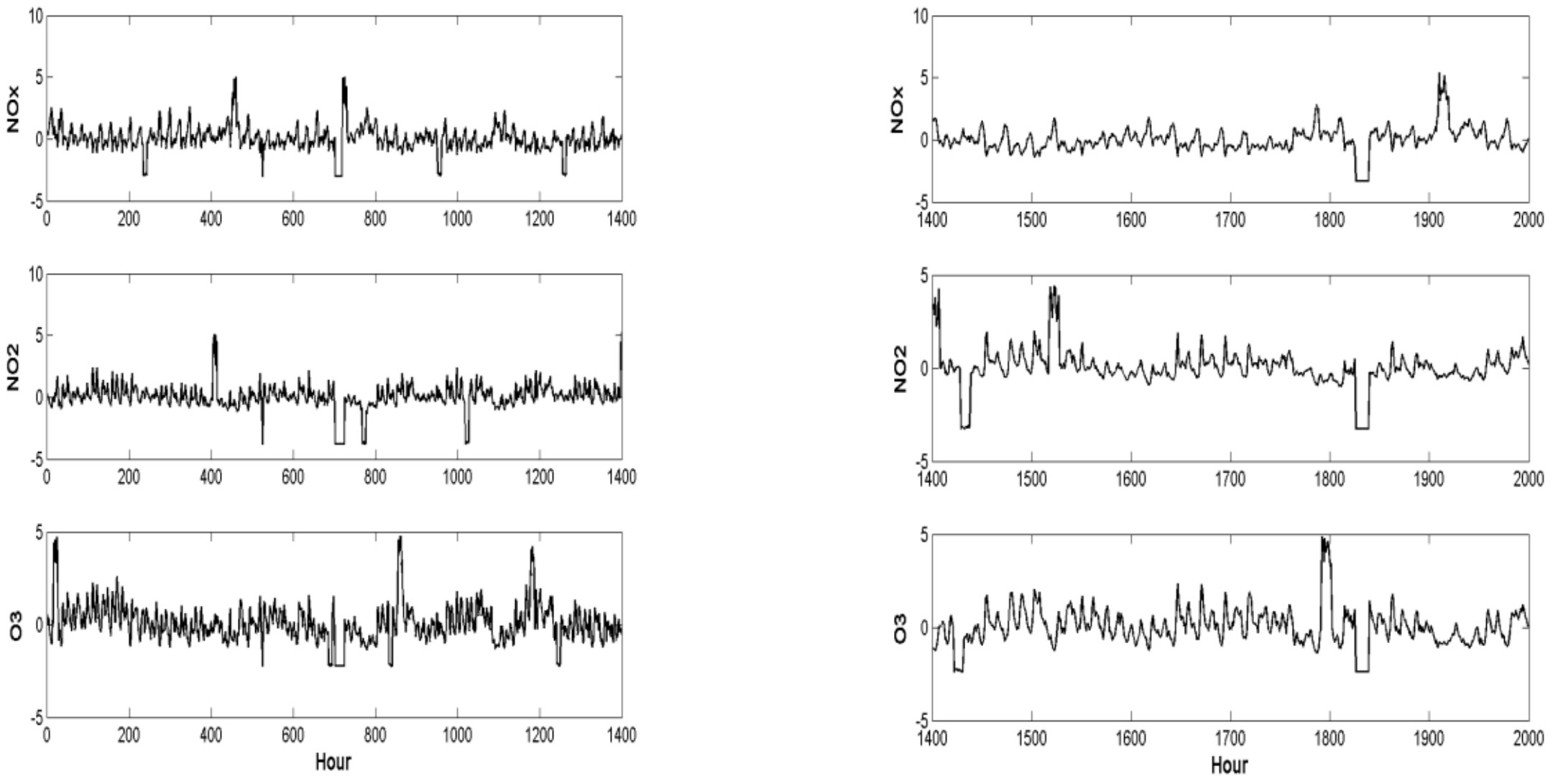

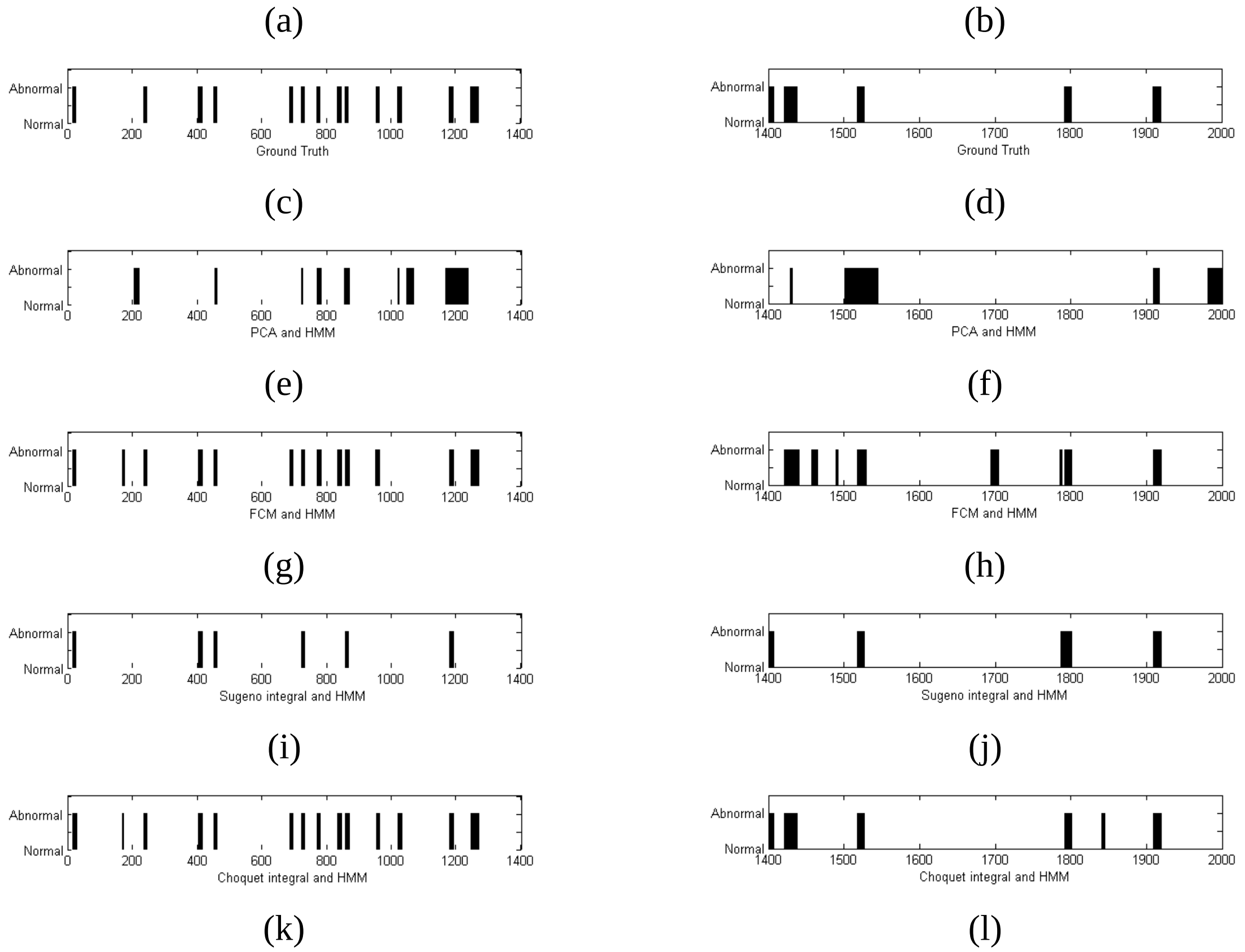

Figure 8 Air Quality Dataset: (a) training set, (b) test set, (c) Ground truth of training set, (d) Ground truth of testing set, (e) Experimental results of PCA + HMM (training set), (f) Experimental results of PCA+HMM (testing set), (g) Experimental results of FCM + HMM (training set), (h) Experimental results of FCM + HMM (testing set), (i) Experimental results of Sugeno integral + HMM (training set), (j) Experimental results of Sugeno integral + HMM (testing set), (k) Experimental results of Choquet integral + HMM (training set), (l) Experimental results of Choquet integral + HMM (testing set).

Table 5 Experimental results obtained for Air Quality Dataset

| Training Set | Accuracy | Sensitivity | Specificity | F-measure |
|---|---|---|---|---|
| PCA+HMM | 0.8586 | 0.9220 | 0.3007 | 0.9214 |
| FCM+HMM | 0.9807 | 0.9865 | 0.9301 | 0.9892 |
| Sugeno_Integral+HMM | 0.9429 | 1.0000 | 0.4406 | 0.9692 |
| Choquet_Integral+HMM | 0.9971 | 0.9968 | 1.0000 | 0.9984 |
| Testing Set | Accuracy | Sensitivity | Specificity | F-measure |
| PCA+HMM | 0.8567 | 0.9048 | 0.3704 | 0.92 |
| FCM+HMM | 0.9367 | 0.9469 | 0.8333 | 0.9646 |
| Sugeno_Integral+HMM | 0.9633 | 0.9908 | 0.6852 | 0.98 |
| Choquet_Integral+HMM | 0.9917 | 0.9908 | 1.0000 | 0.9954 |

As illustrated in the figures and tables, since fuzzy integral and FCM has been used to combine multivariate time series, the performance improvement of the corresponding detectors is quite apparent. This is related with the fact that more useful information is contained in the transformation to an observed sequence. Similar to the experimental results of synthetic dataset, fuzzification coefficient and number of clusters (or observed states) are also associated with the performance of these detectors.

Table 6 Improvement of the proposed detectors vis-à-vis the basic detector with PCA (%)

| | FCM+HMM | Sugeno_Integral+HMM | Choquet_Integral+HMM |
|---|---|---|---|
| U.S. Dollar Exchange Rate Dataset | 6.6474 | 21.0983 | 20.5925 |
| Air Quality Dataset | 14.2263 | 9.8170 | 16.1398 |
| EEG Eye State Dataset | 0.0635 | 5.0810 | 5.3350 |

Table 6 summarizes the improvement of the proposed detectors vis-à-vis the basic detector with PCA when the optimal parameters have been utilized. Compared to the results obtained by applying the PCA to the multivariate time series to form a univariate time series through a linear transform, fuzzy integral is more flexible as the relative importance of different variables is also considered. In summary, the improvement of the ability of detecting anomalies can be attributed to them containing more useful information in the transformation, which might provide more help for HMM-based detectors.

## 5. Conclusions

In this paper, we have investigated the multivariate time series anomaly detection problem by involving different transformation methods and HMM. The objective of this study was to compare different transformation approaches in HMM-based anomaly detection methods. Fuzzy integral and FCM clustering methods can retain more useful information in the transformation process and offer more help for HMM based detectors to deliver better performance. A series of experiments involving synthetic and real dataset is completed to demonstrate the performance of the proposed detectors. Although the proposed anomaly detectors show good performance, there is a major limitation of intensive computing, especially in case of fuzzy integral based detectors. To overcome this problem, a certain alternative would be to engage experts in specifying some initial values of degrees of importance. The method comes with some limitations as we have only concentrated on amplitude anomalies in multivariate time series. Therefore, detecting other types anomalies (e.g., shape anomalies) for larger datasets is a useful further direction. Another pursuit worth investigating is to quantify information loss when transforming from multivariate time series to univariate time series.

**Acknowledgments**

Support from the NSERC CRD project and the Canada Research Chair (CRC) Program is gratefully appreciated.

## 6. References

[1] S.-M. Chen and S.-W. Chen, "Fuzzy forecasting based on two-factors second-order fuzzy-trend logical relationship groups and the probabilities of trends of fuzzy logical relationships," *IEEE Transactions on Cybernetics* vol. 45, no. 3, pp. 391-403, 2015.

[2] G. C. Reinsel, *Elements of multivariate time series analysis*. New York: Springer 2003.

[3] S.-H. Cheng, S.-M. Chen, and W.-S. Jian, "Fuzzy time series forecasting based on fuzzy logical relationships and similarity measures," *Information Sciences,* vol. 327, pp. 272-287, 2016.

[4] S. Lahmiri, "A variational mode decompoisition approach for analysis and forecasting of economic and financial time series," *Expert Systems with Applications,* vol. 55, pp. 268-273, 2016.

[5] R. Rosas-Romero, A. Díaz-Torres, and G. Etcheverry, "Forecasting of stock return prices with sparse representation of financial time series over redundant dictionaries," *Expert Systems with Applications,* vol. 57, pp. 37-48, 2016.

[6] X. Gong, Y.-W. Si, S. Fong, and R. P. Biuk-Aghai, "Financial time series pattern matching with extended UCR Suite and Support Vector Machine," *Expert Systems with Applications,* vol. 55, pp. 284-296, 2016.

[7] L. Maciel, R. Ballini, and F. Gomide, "Evolving granular analytics for interval time series forecasting," *Granular Computing,* vol. 1, no. N/A, pp. 1-12, 2016.

[8] L. Wang, Z. Wang, and S. Liu, "An effective multivariate time series classification approach using echo state network and adaptive differential evolution algorithm," *Expert Systems with Applications,* vol. 43, pp. 237-249, 2016.

[9] G. Heydari, M. Vali, and A. A. Gharaveisi, "Chaotic time series prediction via artificial neural square fuzzy inference system," *Expert Systems with Applications,* vol. 55, pp. 461-468, 2016.

[10] A. Heydari, M. Tavakoli, and N. Salim, "Detection of fake opinions using time series," *Expert Systems with Applications*, vol. 58, pp. 83-92, 2016.

[11] P. Lingras, F. Haider, and M. Triff, "Granular meta-clustering based on hierarchical, network, and temporal connections," *Granular Computing*, vol. 1, no. 1, pp. 71-92, 2016.

[12] H. Cheng, P.-N. Tan, C. Potter, and S. A. Klooster, "Detection and Characterization of Anomalies in Multivariate Time Series," in *Proceedings of the SIAM International Conference on Data Mining (SDM)*, Sparks, Nevada, 2009, vol. 9, pp. 413-424: SIAM.

[13] D. Zheng, F. Li, and T. Zhao, "Self-adaptive statistical process control for anomaly detection in time series," *Expert Systems with Applications*, vol. 57, pp. 324-336, 2016.

[14] B. R. Helliker and S. L. Richter, "Subtropical to boreal convergence of tree-leaf temperatures," *Nature*, vol. 454, no. 7203, pp. 511-514, 2008.

[15] M.-H. Chen, W. G. Kim, and H. J. Kim, "The impact of macroeconomic and non-macroeconomic forces on hotel stock returns," *International Journal of Hospitality Management*, vol. 24, no. 2, pp. 243-258, 2005.

[16] M. A. Hayes, "Contextual Anomaly Detection Framework for Big Sensor Data," Thesis, The University of Western Ontario, Ontario, Canada, 2014.

[17] N. Takeishi and T. Yairi, "Anomaly detection from multivariate time-series with sparse representation," in *Systems, Man and Cybernetics (SMC), 2014 IEEE International Conference on*, San Diego, USA, 2014, pp. 2651-2656: IEEE.

[18] T. Idé, S. Papadimitriou, and M. Vlachos, "Computing correlation anomaly scores using stochastic nearest neighbors," in *Proceedings of the 7th IEEE International Conference on Data Mining (ICDM),* Omaha, USA, 2007, pp. 523-528: IEEE.

[19] H. Qiu, Y. Liu, N. A. Subrahmanya, and W. Li, "Granger causality for time-series anomaly detection," in *Proceedings of the 12th IEEE International Conference on Data Mining (ICDM)*, Brussels, Belgium, 2012, pp. 1074-1079: IEEE.

[20] M. Gan, C. P. Chen, H.-X. Li, and L. Chen, "Gradient radial basis function based varying-coefficient autoregressive model for nonlinear and nonstationary time series," *IEEE Signal Processing Letters*, vol. 22, no. 7, pp. 809-812, 2015.

[21] K. Bisht and S. Kumar, "Fuzzy time series forecasting method based on hesitant fuzzy sets," *Expert Systems with Applications,* vol. N/A, no. N/A, p. N/A, 2016.

[22] M. Jones, D. Nikovski, M. Imamura, and T. Hirata, "Exemplar learning for extremely efficient anomaly detection in real-valued time series," *Data Mining and Knowledge Discovery*, vol. N/A, no. N/A, pp. 1-28, 2016.

[23] Z. Xu and X. Gou, "An overview of interval-valued intuitionistic fuzzy information aggregations and applications," *Granular Computing*, vol. 1, no. N/A, pp. 1-27, 2016.

[24] I. Ben-Gal, "Outlier detection," in Data mining and knowledge discovery handbook: Springer, 2005, pp. 131-146.

[25] P. Filzmoser, "A multivariate outlier detection method," in *Proceedings of the Seventh International Conference on Computer Data Analysis and Modeling*, Minsk, Belarus, 2004, pp. 18-22.

[26] L. R. Rabiner, "A tutorial on Hidden Markov Models and selected applications in speech recognition," *Proceedings of the IEEE,* vol. 77, no. 2, pp. 257-286, 1989.

[27] R. J. Elliott, T. K. Siu, and E. S. Fung, "A Double HMM approach to Altman Z-scores and credit ratings," *Expert Systems with Applications,* vol. 41, no. 4, pp. 1553-1560, 2014.

[28] Z. Li, H. Fang, and M. Huang, "Diversified learning for continuous Hidden Markov Models with application to fault diagnosis," *Expert Systems with Applications,* vol. 42, no. 23, pp. 9165-9173, 2015.

[29] R. Fu, H. Wang, and W. Zhao, "Dynamic driver fatigue detection using Hidden Markov Model in real driving condition," *Expert Systems with Applications,* vol. 63, pp. 397-411, 2016.

[30] Y. Cao, Y. Li, S. Coleman, A. Belatreche, and T. M. McGinnity, "Adaptive hidden Markov model with anomaly states for price manipulation detection," *IEEE Transactions on Neural Networks and Learning Systems,* vol. 26, no. 2, pp. 318-330, 2015.

[31] A. Soualhi, G. Clerc, H. Razik, and F. Guillet, "Hidden Markov Models for the prediction of impending faults," *IEEE Transactions on Industrial Electronics,* vol. 63, no. 5, pp. 3271-3281, 2016.

[32] G. D. Forney, "The Viterbi algorithm," *Proceedings of the IEEE,* vol. 61, no. 3, pp. 268-278, 1973.

[33] L. Chen, C. P. Chen, and M. Lu, "A multiple-kernel Fuzzy C-means algorithm for image segmentation," *IEEE Transactions on Systems, Man, and Cybernetics, Part B (Cybernetics),* vol. 41, no. 5, pp. 1263-1274, 2011.

[34] N. R. Pal and J. C. Bezdek, "On cluster validity for the Fuzzy C-means model," *IEEE Transactions on Fuzzy systems,* vol. 3, no. 3, pp. 370-379, 1995.

[35] J. Zhou, C. P. Chen, L. Chen, and H.-X. Li, "A collaborative fuzzy clustering algorithm in distributed network environments," *IEEE Transactions on Fuzzy Systems,* vol. 22, no. 6, pp. 1443-1456, 2014.

[36] H. Tahani and J. M. Keller, "Information fusion in computer vision using the fuzzy integral," *IEEE Transactions on Systems, Man, and Cybernetics,* vol. 20, no. 3, pp. 733-741, 1990.

[37] J. Wu, F. Chen, C. Nie, and Q. Zhang, "Intuitionistic fuzzy-valued Choquet integral and its application in multicriteria decision making," *Information Sciences,* vol. 222, pp. 509-527, 2013.

[38] C.-M. Hwang, M.-S. Yang, W.-L. Hung, and M.-G. Lee, "A similarity measure of intuitionistic fuzzy sets based on the Sugeno integral with its application to pattern recognition," *Information Sciences,* vol. 189, pp. 93-109, 2012.

[39] J. J. Liou, Y.-C. Chuang, and G.-H. Tzeng, "A fuzzy integral-based model for supplier evaluation and improvement," *Information Sciences,* vol. 266, pp. 199-217, 2014.

[40] B.-S. Yoo and J.-H. Kim, "Fuzzy integral-based gaze control of a robotic head for human robot interaction," *IEEE Transactions on Cybernetics* vol. 45, no. 9, pp. 1769-1783, 2015.

[41] A. C. B. Abdallah, H. Frigui, and P. Gader, "Adaptive local fusion with fuzzy integrals," *IEEE Transactions on Fuzzy Systems,* vol. 20, no. 5, pp. 849-864, 2012.

[42] T. Murofushi and M. Sugeno, "An interpretation of fuzzy measures and the Choquet integral as an integral with respect to a fuzzy measure," *Fuzzy sets and Systems,* vol. 29, no. 2, pp. 201-227, 1989.

[43] T. Onisawa, M. Sugeno, Y. Nishiwaki, H. Kawai, and Y. Harima, "Fuzzy measure analysis of public attitude towards the use of nuclear energy," *Fuzzy sets and systems,* vol. 20, no. 3, pp. 259-289, 1986.

[44] H.-C. Wang, C.-S. Lee, and T.-H. Ho, "Combining subjective and objective QoS factors for personalized web service selection," *Expert Systems with Applications*, vol. 32, no. 2, pp. 571-584, 2007.

[45] Y. Chi and T. Zhang, "Study on optimum fusion algorithms of IKONOS high spatial resolution remote sensing image," in *Proceedings of the International Conference on Multimedia Technology (ICMT)*, Hangzhou, China, 2011, pp. 761-764: IEEE.

[46] (2003, Jul.). DataMarket - Time Series Data Library . [Online]. Available: http://robjhyndman.com/tsdldata

[47] (2007, May.). UCI Machine Learning Repository. [Online]. Available: http://www.ics.uci.edu/~mlearn/MLRepository.html